\documentclass[preprint,12pt]{elsarticle}

\usepackage{amsmath,amsfonts,amssymb}
\usepackage{graphicx}
\usepackage{setspace}
\usepackage{tocloft}
\usepackage{textcomp}
\usepackage{color,soul}
\usepackage{subcaption}
\usepackage{float}
\usepackage[colorlinks=true, allcolors=blue]{hyperref}
\usepackage[table,xcdraw,dvipsnames]{xcolor}
\usepackage{lineno}
\usepackage{array}
\usepackage{listings}
\usepackage{mathptmx}
\usepackage{epsfig}
\usepackage{rotating}
\usepackage{makeidx}
\usepackage{url}
\usepackage{xurl}
\usepackage{booktabs}
\usepackage{tabularx}
\usepackage{blindtext}
\usepackage{times}
\usepackage{nccmath}
\usepackage{mwe}
\usepackage{acro}
\usepackage{adjustbox}
\usepackage{colortbl}
\usepackage{relsize}
\usepackage{pifont}
\usepackage{multirow}
\usepackage{multicol}
\usepackage{nicematrix}
\usepackage{bm}
\usepackage{makecell}
\usepackage{tabu}
\usepackage[capitalize]{cleveref}
\DeclareMathSymbol{\shortminus}{\mathbin}{AMSa}{"39}

\usepackage{color}
\definecolor{dkgreen}{rgb}{0,0.6,0}
\definecolor{gray}{rgb}{0.5,0.5,0.5}
\definecolor{mauve}{rgb}{0.58,0,0.82}

\colorlet{Mycolor1}{green!10!orange}

\usepackage{lineno}

\journal{Magnetic Resonance Imaging}

\begin{document}

\begin{frontmatter}



\title{OpenMASC: An Open-Source Pipeline for Cross-Trajectory Metal-Aware Sampling and Correction in Accelerated MRI}

\author[1]{Zhengyi Lu}
\author[2]{Ming Lu}
\author[1]{Chongyu Qu}
\author[1]{Junchao Zhu}
\author[1]{Junlin Guo}
\author[1]{Marilyn Lionts}
\author[1]{Yanfan Zhu}
\author[1]{Yuechen Yang}
\author[1]{Tianyuan Yao}
\author[3]{Jayasai Rajagopal}
\author[1]{Bennett Allan Landman}
\author[3]{Xiao Wang}
\author[2]{Xinqiang Yan}
\author[1]{Yuankai Huo}
\affiliation[1]{organization={Vanderbilt University},
            city={Nashville},
            state={TN},
            country={USA}}
\affiliation[2]{organization={Vanderbilt University Medical Center},
            city={Nashville},
            state={TN},
            country={USA}}
\affiliation[3]{organization={Oak Ridge National Laboratory},
            city={Oak Ridge},
            state={TN},
            country={USA}}

\begin{abstract}
Metal implants corrupt MRI measurements throughout $k$-space, yet existing accelerated MRI methods assume clean data and most metal artifact reduction approaches assume fully sampled acquisitions. No public dataset provides paired $k$-space and images with and without metal for the same anatomy, and no framework jointly addresses artifact-aware acquisition and reconstruction across sampling trajectories. We present OpenMASC, an open-source pipeline covering the full workflow from data generation to deployment. A physics-based data generation module converts public CT volumes into paired clean and metal-corrupted MRI data in both Cartesian and radial formats. MA-VarNet, an unrolled reconstruction network with a per-cascade DC Rectifier, corrects artifacts that data-consistency steps reintroduce from corrupted measurements. A reinforcement learning agent actively selects $k$-space readouts and co-trains with the reconstruction network through a decoupled three-stage procedure. The framework is trajectory-agnostic except for the data-consistency operator, supporting both Cartesian and radial acquisition without architectural changes. Experiments on two datasets at $4\times$ and $8\times$ acceleration demonstrate consistent improvements over conventional and learned baselines on both trajectories.

\end{abstract}


\end{frontmatter}

\section{Introduction}
Millions of patients carry metallic implants such as hip prostheses and spinal hardware~\cite{kremers2015prevalence, hegde2023highlights}, and MRI is frequently needed to assess the tissue surrounding these devices. However, the metal distorts the magnetic field and corrupts the acquired $k$-space data, producing artifacts that can obscure critical anatomy~\cite{hargreaves2011metal, koch2010magnetic}. Separately, MRI is inherently slow, and accelerated methods that acquire only a subset of $k$-space have become a major research focus~\cite{lin2000principles, moratal2008k, knoll2020fastmri}. These two challenges have largely been addressed in isolation: existing acceleration methods assume clean measurements, while deep learning approaches for MAR assume fully sampled acquisitions. When both conditions occur together, as they do for any patient with an implant who needs a fast scan, neither line of work applies directly. A key barrier is also the absence of any public dataset providing paired $k$-space and reconstructed images with and without metal for the same anatomy, which is required for both supervised MAR training and RL-based acquisition policy learning.
 
The joint problem also raises architectural and design challenges. An RL sampling agent trained on artifact-free data learns acquisition patterns that become ineffective under metal corruption~\cite{pineda2020active, yen2024adaptive, xu2025reinforcement}. Modern unrolled reconstruction networks alternate between learned refinement and DC steps that enforce agreement with the acquired $k$-space~\cite{sriram2020end, ramzi2020xpdnet, giannakopoulos2024accelerated}. When measurements are corrupted, DC reintroduces the very artifacts the refinement step removed, and this cycle worsens with each cascade. On top of this, most existing methods support only Cartesian sampling, yet radial trajectories are often preferred near metal for their robustness to motion. Current pipelines cannot be transferred across these geometries without substantial redesign.

\begin{figure*}[t]
\begin{center}
\includegraphics[width=1\linewidth]{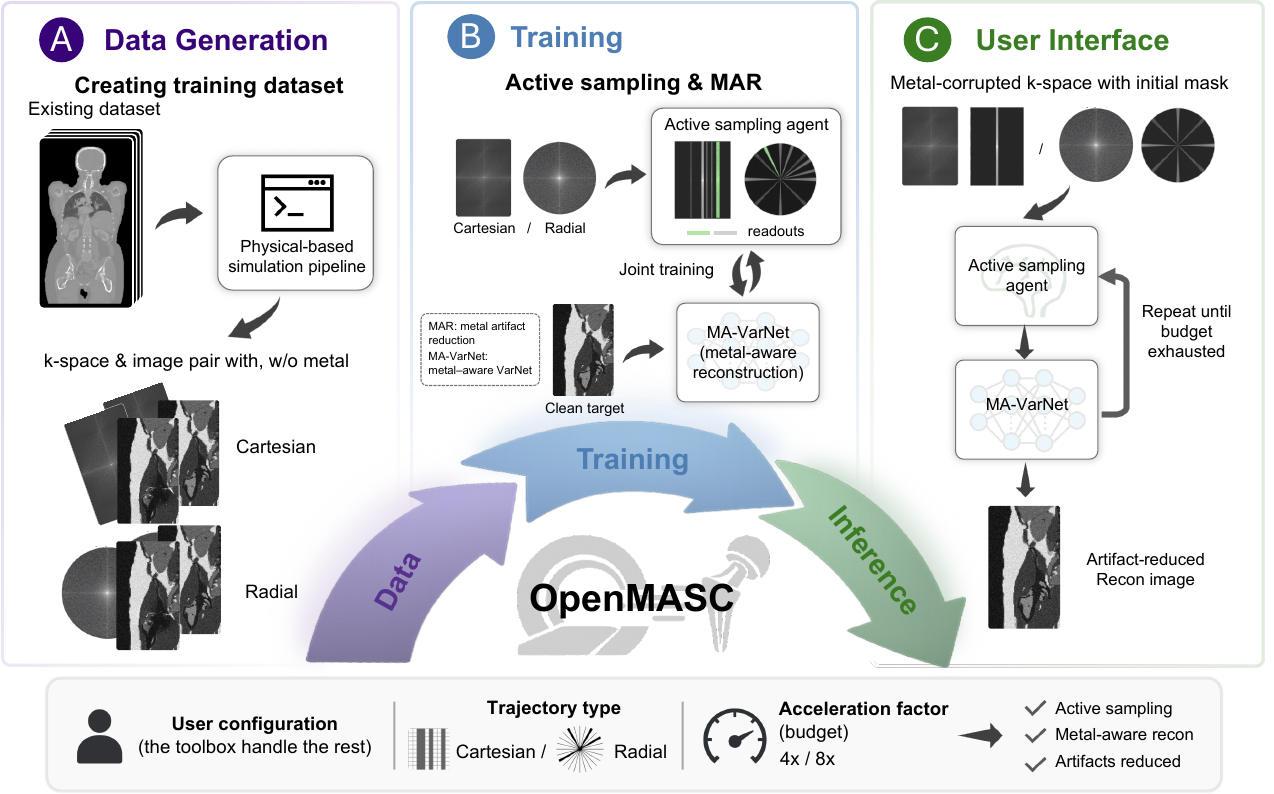}
\end{center}
\caption{Overview of the OpenMASC pipeline. The user selects a trajectory type (Cartesian or radial) and an acceleration factor ($4\times$ or $8\times$); the toolbox handles the rest. \textbf{(A)}~The Data Generation module converts existing CT volumes into paired clean and metal-corrupted MRI data via physics-based simulation, in either Cartesian or radial format. \textbf{(B)}~An active sampling agent and MA-VarNet (metal-aware reconstruction) are jointly trained on the generated data. \textbf{(C)}~At deployment, the agent iteratively selects $k$-space readouts while MA-VarNet reconstructs, repeating until the acquisition budget is exhausted to produce an artifact-reduced image.}
\label{fig:overview}
\end{figure*}

In this work, we present OpenMASC, an open-source pipeline that spans the full workflow from training data generation to cross-trajectory metal-aware accelerated MRI reconstruction. As illustrated in Fig.~\ref{fig:overview}, OpenMASC is organized into three modules. The user specifies only two configuration choices, trajectory type (Cartesian or radial) and acceleration factor ($4\times$ or $8\times$), and the toolbox handles the rest. A physics-based \textbf{Data Generation} module (Fig.~\ref{fig:overview}A) converts existing CT volumes into paired clean and metal-corrupted MRI data in either trajectory format. The \textbf{Active Sampling and MAR module} (Fig.~\ref{fig:overview}B) jointly trains an active sampling agent and MA-VarNet, a metal-aware reconstruction network, so that the acquisition policy and artifact correction co-adapt. At deployment, the \textbf{User Interface} (Fig.~\ref{fig:overview}C) iteratively acquires readouts and reconstructs until the budget is exhausted, producing an artifact-reduced image. Crucially, the DC operator is the sole trajectory-specific component in the entire framework; all other modules are shared across geometries without modification.
 
Our contributions can be summarized as follows:
\begin{itemize}
    \item An end-to-end open-source pipeline for metal-aware accelerated MRI covering data simulation, model training, and deployment, configurable by trajectory type and acceleration factor (Fig.~\ref{fig:overview}).
    \item A physics-based data generation module that generates paired clean and metal-corrupted MRI volumes from existing CT datasets in both Cartesian and radial $k$-space (Fig.~\ref{fig:overview}A).
    \item MA-VarNet, an unrolled reconstruction network with per-cascade DC Rectifier that breaks the artifact reintroduction cycle caused by data consistency under metal corruption (Fig.~\ref{fig:overview}B).
    \item A jointly trained active sampling agent that co-adapts with the reconstruction network across trajectory geometries (Fig.~\ref{fig:overview}B).
    \item Comprehensive experiments on two datasets demonstrating consistent improvements over conventional and learned baselines, with ablations confirming the necessity of each component.
\end{itemize}

\section{Related Work}
\label{sec:related_work}

\subsection{Metal Artifact Reduction in MRI}
\label{sec:rw_mar}

The susceptibility difference between metallic implants and surrounding tissue induces spatially varying off-resonance fields that disrupt the MRI encoding process, leading to signal voids, geometric warping, and pile-up artifacts~\cite{hargreaves2011metal,koch2010magnetic}. Because these phase errors are spatially distributed, their effects propagate across the entire $k$-space rather than remaining localized near the implant~\cite{schenck1996role,haskell2023off}.

Conventional remedies operate at the pulse-sequence level. View angle tilting reduces through-plane distortion by applying a compensating gradient during readout. SEMAC and MAVRIC resolve off-resonance along the slice direction through additional phase encodes, at the cost of significantly longer scan times~\cite{feuerriegel2024managing}. Seo et al.~\cite{seo2020artificial} applied neural networks to accelerate SEMAC-based correction.

On the algorithmic side, deep learning has been applied to suppress metal artifacts in the image domain. Kwon et al.~\cite{kwon2018learning} trained a network on simulated off-resonance data to correct artifacts from dual-polarity readout acquisitions. A common limitation of these approaches is their reliance on fully sampled input: they are designed to post-process complete reconstructions and do not account for the aliasing patterns that arise under accelerated sampling. Integrating artifact correction into the reconstruction pipeline itself remains an open challenge.

\subsection{Accelerated MRI Reconstruction and Active Acquisition}
\label{sec:rw_accel}

Reducing MRI scan time by collecting fewer $k$-space samples has been pursued through both classical and learning-based reconstruction. Deep unrolled networks have become the leading approach, mapping iterative optimization into end-to-end trainable cascades where learned modules alternate with DC steps. Sriram et al.~\cite{sriram2020end} proposed E2E-VarNet, a variational network with end-to-end training for accelerated reconstruction. Giannakopoulos et al.~\cite{giannakopoulos2024accelerated} extended this with FI-VarNet, a dual-path architecture operating in both feature and image domains. Other unrolled designs include XPDNet~\cite{ramzi2020xpdnet}. These networks implicitly trust the acquired data: the DC step projects the reconstruction toward the raw measurements at every cascade. Under metal corruption, this trust is misplaced, as DC forces agreement with distorted observations and counteracts the refinement block, an issue we address with our DC Rectifier module.

Which $k$-space locations to sample is itself an important design choice. Fixed patterns such as equispaced, variable-density, and low-frequency-biased masks are widely used but content-agnostic. Learned acquisition policies cast the problem as sequential decision-making. Pineda et al.~\cite{pineda2020active} formulated active $k$-space sampling with RL using a subject-specific Double DQN agent. Bakker et al.~\cite{bakker2020experimental} proposed greedy policy search for MRI experimental design. Zhang et al.~\cite{zhang2019reducing} addressed uncertainty reduction in undersampled reconstruction through active acquisition. Yen et al.~\cite{yen2024adaptive} explored adaptive $k$-space sampling for rapid pathology prediction. Policy-gradient methods based on PPO~\cite{schulman2017proximal} have also been applied, scaling more naturally to large action spaces. A shared assumption across all these methods is that collected samples faithfully represent the underlying signal. None account for measurement-level corruption, and all operate on Cartesian grids.

Radial sampling provides complementary strengths for clinical metal imaging. Each spoke passes through the $k$-space center, providing natural oversampling of low frequencies and inherent motion tolerance~\cite{block2014towards}. Golden-angle radial sampling~\cite{feng2014golden} enables flexible retrospective reconstruction. Reconstruction requires the NUFFT~\cite{fessler2003nonuniform} and appropriate density compensation, with libraries such as MRI-NUFFT~\cite{comby2025mri} simplifying implementation. Despite these advantages, learned acquisition and artifact correction have not been extended to radial trajectories, and no existing framework supports both geometries within a single architecture.

\subsection{MRI Simulation and Paired Dataset Generation}
\label{sec:rw_data}

Training the components described above, both the MAR network and the RL sampling agent, requires access to exactly matched $k$-space and image pairs acquired with and without metal. Public MRI repositories such as fastMRI~\cite{zbontar2018fastmri,knoll2020fastmri} provide large-scale clinical data but contain no metal-corrupted volumes. Post-hoc augmentation with simplified artifact models~\cite{kwon2018learning} can approximate corruption in the image domain, yet does not produce physically accurate $k$-space-level distortions tied to a known implant geometry.

Physics-based simulation offers a path forward. An open-source MRI simulator~\cite{keskin2026open} can model the full signal chain, including susceptibility-induced field inhomogeneities, signal dephasing, and sequence-specific encoding, to produce matched clean and corrupted $k$-space. However, this simulator relies on detailed, manually segmented human phantoms that are expensive to create and cannot be scaled to large numbers of diverse subjects. No existing pipeline automates paired data generation from public datasets or supports producing both Cartesian and radial $k$-space from the same underlying anatomy.

\section{Method}
\label{sec:method}

\begin{figure*}[t]
    \centering
    \includegraphics[width=\textwidth]{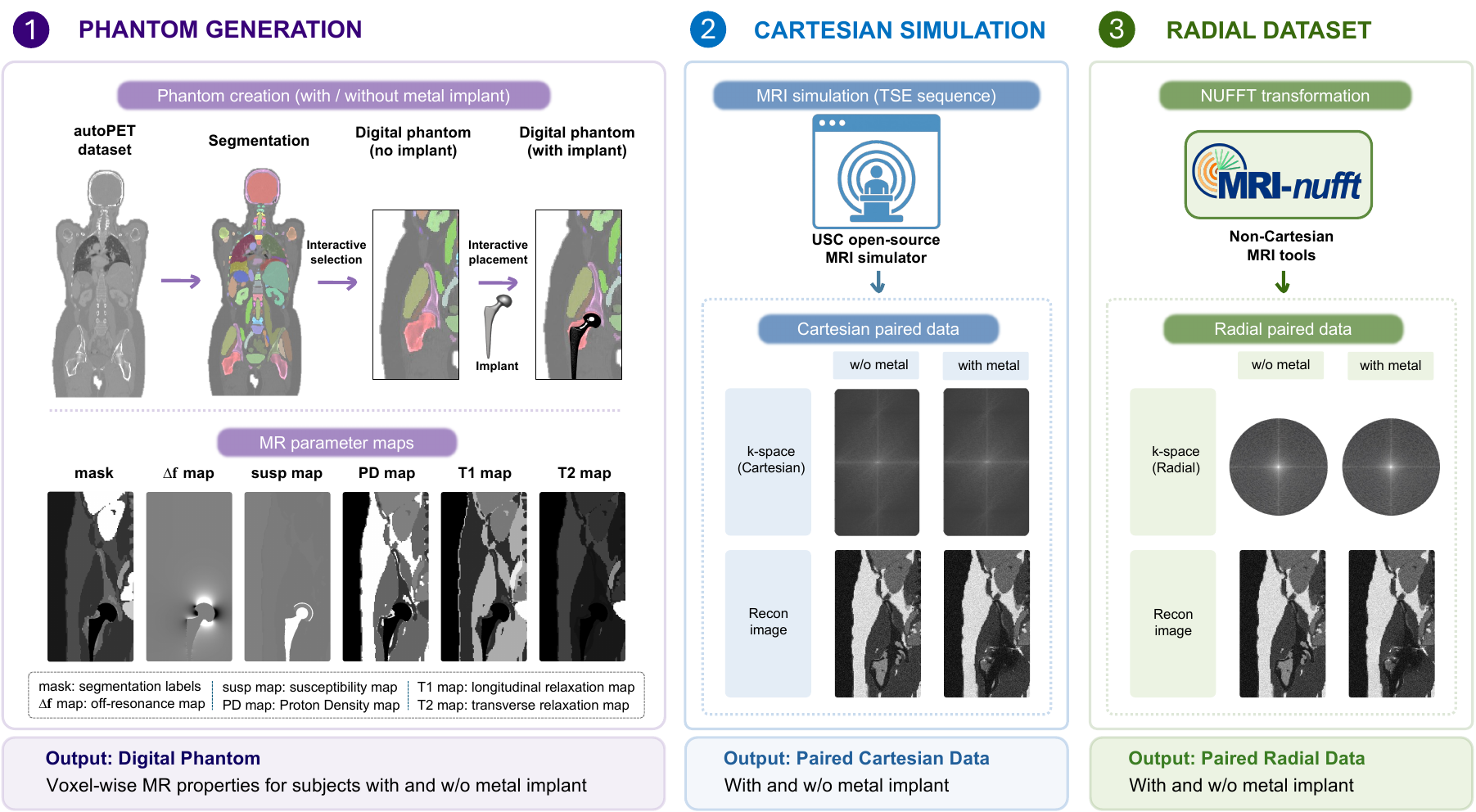}
    \caption{Data Generation pipeline. \textbf{Stage~1:} CT volumes from the autoPET dataset are automatically segmented into multi-tissue phantoms. The user interactively selects the implant site and positions the virtual prosthesis, producing paired digital phantoms (with and without metal) and six voxel-wise MR property maps. \textbf{Stage~2:} An open-source MRI simulator generates paired Cartesian $k$-space and reconstructed images via a TSE sequence. \textbf{Stage~3:} Radial $k$-space and images are derived from the same paired data via NUFFT using MRI-NUFFT.}
    \label{fig:data_generation}
\end{figure*}

OpenMASC comprises three modules (Fig.~\ref{fig:overview}): a Data Generation module that produces paired training data from existing CT scans (Sec.~\ref{sec:data_generation}), a joint active sampling and reconstruction framework built on MA-VarNet (Secs.~\ref{sec:problem}--\ref{sec:mavarnet}), and a three-stage training pipeline that enables stable co-adaptation of all learnable components (Sec.~\ref{sec:training}).

\subsection{Data Generation}
\label{sec:data_generation}

Supervised training for both the reconstruction network and the acquisition agent requires paired $k$-space and image data with and without metal artifacts for the same anatomy. As no such public dataset exists, we build a physics-based simulation pipeline that generates this data automatically from existing CT scans. The Data Generation module, shown in Fig.~\ref{fig:data_generation}, proceeds in three stages: phantom generation, Cartesian simulation, and radial dataset creation.

\paragraph{Stage 1: Phantom generation} The pipeline accepts CT volumes from a public repository (in our case, the autoPET PET/CT dataset~\cite{gatidis2023autopet}) and converts them into digital phantoms with voxel-wise MR tissue properties. Segmentation is performed by TotalSegmentator~\cite{wasserthal2023totalsegmentator} with three complementary task configurations: (1) a general task that delineates individual organs, (2) a tissue-composition task that separates fat from muscle, and (3) a body-region task for anatomical localization. These label maps are fused into a unified multi-tissue phantom, and literature-based MR parameters at 3T~\cite{bottomley1987review,pohmann2016signal,rooney2007magnetic,bojorquez2017normal,gold2004musculoskeletal,stanisz2005t1} are assigned per tissue class, yielding maps of proton density (PD), longitudinal relaxation time ($T_1$), and transverse relaxation time ($T_2$).

To generate the metal-corrupted counterpart, the pipeline provides an interactive interface for selecting the implant site and positioning a virtual prosthesis. We use a total hip arthroplasty model consisting of a femoral stem, femoral head, acetabular liner, and acetabular cup~\cite{modinger2023magnetic}. The implant material is cobalt-chromium with a susceptibility of 900\,ppm; surrounding tissues are assigned default susceptibility values of $-9.05$\,ppm (soft tissue and water), $-8.86$\,ppm (cortical bone), and $-5.55$\,ppm (fat)~\cite{schenck1996role,smith2015characterizing}. The result is a pair of phantoms per subject, one without and one with the implant, together with six property maps: segmentation mask, off-resonance ($\Delta f$) map, susceptibility map, PD map, $T_1$ map, and $T_2$ map (Fig.~\ref{fig:data_generation}, Stage~1).

\paragraph{Stage 2: Cartesian simulation} Each phantom pair is fed into an open-source MRI simulator~\cite{keskin2026open} that reproduces the full signal formation chain at 3T, including susceptibility-driven field inhomogeneity, intravoxel dephasing, and geometric distortion. The simulator runs a turbo spin echo (TSE) protocol with parameters chosen to match a typical clinical hip scan: TR\,=\,4050\,ms, TE\,=\,32\,ms, readout bandwidth\,=\,710\,Hz/pixel, RF bandwidth\,=\,1\,kHz, and slice thickness\,=\,3\,mm. The output for each subject comprises (1) artifact-free $k$-space and its reconstruction, (2) metal-corrupted $k$-space and its reconstruction, and (3) a binary implant mask. Cartesian $k$-space has matrix dimensions of $400 \times 200$ (readout $\times$ phase-encode), obtained via the 2D DFT of the complex image. Each corrupted volume is exactly paired with its clean counterpart from the same anatomy, supplying both the supervision signal for MAR training and the reference for RL reward computation (Fig.~\ref{fig:data_generation}, Stage~2).

\paragraph{Stage 3: Radial dataset creation} The same paired complex images serve as the starting point for generating non-Cartesian data. The pipeline resamples each image onto 200 center-out radial spokes with 400 points per spoke through the NUFFT, using the MRI-NUFFT library~\cite{comby2025mri}. Adjoint NUFFT with density compensation produces the corresponding radial reconstructions. Because both trajectory formats originate from identical phantom pairs, differences in downstream reconstruction quality can be attributed to the sampling geometry itself rather than to anatomical variation (Fig.~\ref{fig:data_generation}, Stage~3).

\subsection{Problem Formulation and Active Acquisition}
\label{sec:problem}

For single-coil MRI, the relationship between an image $\mathbf{x} \in \mathbb{C}^{H \times W}$ and its $k$-space representation $\mathbf{y}$ is given by $\mathbf{y} = \mathcal{F}\{\mathbf{x}\}$, where $\mathcal{F}$ denotes the DFT for Cartesian grids or the NUFFT~\cite{fessler2003nonuniform} for radial trajectories. When a metallic implant is present, $B_0$ field inhomogeneity introduces a spatially dependent phase error $\phi(\mathbf{r})$ that corrupts every measurement:
\begin{equation}
    \mathbf{y}_{\mathrm{metal}} = \mathcal{F}\{\mathbf{x} \cdot e^{j\phi(\mathbf{r})}\} + \boldsymbol{\epsilon},
    \label{eq:metal_corruption}
\end{equation}
with $\boldsymbol{\epsilon}$ representing noise. Given an acquisition budget of $B$ readouts, only a subset is observed: $\mathbf{y}_{\mathrm{sub}} = \mathbf{M} \odot \mathbf{y}_{\mathrm{metal}}$, where $\mathbf{M}$ selects $B$ phase-encode lines (Cartesian) or spokes (radial).

The problem of choosing which readouts to acquire is cast as an MDP~\cite{sutton1998reinforcement}. The state at step $t$ comprises the current magnitude reconstruction and acquisition mask, $\mathbf{s}_t = [|\hat{\mathbf{x}}_t|, \mathbf{M}_t] \in \mathbb{R}^{2 \times H \times W}$. The agent selects an action $a_t$ corresponding to the next readout and receives a reward proportional to the resulting quality improvement:
\begin{equation}
    r_t = \alpha \cdot \bigl(\mathrm{SSIM}(\hat{\mathbf{x}}_{t+1}, \mathbf{x}) - \mathrm{SSIM}(\hat{\mathbf{x}}_t, \mathbf{x})\bigr),
    \label{eq:reward}
\end{equation}
where $\alpha$ scales the signal and $\mathbf{x}$ is the artifact-free ground truth. A convolutional actor-critic network parameterizes the policy~\cite{schulman2017proximal}: the actor head outputs logits over all readouts (previously acquired positions are masked to $-\infty$), and the critic head estimates the state value. Training follows the clipped PPO objective with GAE~\cite{schulman2015high}.

Because the agent operates on reconstructed magnitude images and binary masks rather than on raw $k$-space, its architecture and training are identical for Cartesian and radial settings. Only the interpretation of each action changes: selecting a phase-encode line in one case or a spoke in the other.

\subsection{MA-VarNet: Metal-Aware Reconstruction}
\label{sec:mavarnet}

MA-VarNet builds on the variational network~\cite{sriram2020end} and consists of $T$ cascades. Each cascade $i$ applies a U-Net refinement followed by a DC step and a DCR module. Starting from the zero-filled estimate $\hat{\mathbf{x}}_0 = \mathcal{F}^{-1}\{\mathbf{M} \odot \mathbf{y}\}$, the cascades progressively improve reconstruction quality.

\paragraph{U-Net refinement} A cascade-specific U-Net $g_{\theta_i}$ maps the previous output to a refined estimate:
\begin{equation}
    \tilde{\mathbf{x}}_i = g_{\theta_i}(\hat{\mathbf{x}}_{i-1}).
\end{equation}

\paragraph{Data consistency} The refined estimate is then projected toward the acquired measurements. For Cartesian data this takes a closed-form $k$-space blending:
\begin{equation}
    \hat{\mathbf{x}}^{\mathrm{DC}}_i = \mathcal{F}^{-1}\!\bigl\{\frac{\mathbf{M} \odot \mathbf{Y} + \lambda_i \cdot \mathcal{F}\{\tilde{\mathbf{x}}_i\}}{\mathbf{M} + \lambda_i}\bigr\}, \quad \textit{(Cartesian)}
    \label{eq:dc_cartesian}
\end{equation}
while radial geometry uses a gradient-based correction:
\begin{equation}
    \hat{\mathbf{x}}^{\mathrm{DC}}_i = \tilde{\mathbf{x}}_i - \frac{\lambda_i}{\|\mathcal{A}^\mathsf{H} \mathbf{D} \mathcal{A}\|} \cdot \mathcal{A}^\mathsf{H} \mathbf{D} (\mathcal{A}\tilde{\mathbf{x}}_i - \mathbf{y}), \quad \textit{(Radial)}
    \label{eq:dc_radial}
\end{equation}
Here $\lambda_i$ is a learnable scalar per cascade, $\mathcal{A}$ and $\mathcal{A}^\mathsf{H}$ denote the NUFFT forward and adjoint operators~\cite{fessler2003nonuniform}, and $\mathbf{D}$ applies density compensation. This DC block is the sole module that differs between the two trajectories; every other component in the framework remains unchanged.

\paragraph{DC Rectifier} Metal corruption means that all acquired samples carry phase errors (Eq.~\ref{eq:metal_corruption}). The DC step therefore pulls the reconstruction back toward distorted data at every cascade, undoing part of the preceding refinement. Repeated application of this pattern causes artifact energy to grow through the network. The DCR addresses this by applying a residual correction in the magnitude domain immediately after each DC step:
\begin{equation}
    m_i = |\hat{\mathbf{x}}^{\mathrm{DC}}_i|, \quad \bar{m}_i = m_i \,/\, p_{98}(m_i), \quad \hat{m}_i = \mathrm{ReLU}\!\bigl(h_{\psi_i}(\bar{m}_i)\bigr) \cdot p_{98}(m_i),
    \label{eq:dcr}
\end{equation}
where $p_{98}(\cdot)$ is the 98th-percentile value used for stable intensity normalization and the residual network $h_{\psi_i}(\bar{m}_i) = \bar{m}_i + \Delta_{\psi_i}(\bar{m}_i)$ learns to predict an additive correction. The phase from the DC output is preserved when forming the corrected complex image:
\begin{equation}
    \hat{\mathbf{x}}_i = \hat{m}_i \cdot e^{j \arg(\hat{\mathbf{x}}^{\mathrm{DC}}_i)}.
    \label{eq:phase_recompose}
\end{equation}
Operating entirely on magnitudes makes the DCR agnostic to the sampling trajectory. Placing it after every cascade, rather than only at the final output, interrupts the accumulation of DC-induced errors at each stage.

\paragraph{Cross-trajectory design} The entire MA-VarNet architecture is trajectory-agnostic except for one component: the DC block. Cartesian and radial geometries use different forward models (Eqs.~\ref{eq:dc_cartesian} and~\ref{eq:dc_radial}), but the U-Net refinement, DCR, and acquisition agent all operate on magnitude images and binary masks, which carry no trajectory-specific structure. In practice, switching from Cartesian to radial requires only replacing the DC operator (FFT $\leftrightarrow$ NUFFT) and reinterpreting each action as a spoke rather than a phase-encode line. The loss functions, training procedure, and all network weights outside the DC block remain identical.

\subsection{Decoupled Training}
\label{sec:training}

Naively training all components together is unstable: the DC step drives the reconstruction toward corrupted measurements while the loss function rewards agreement with clean targets, producing conflicting gradients. We therefore adopt a decoupled three-stage procedure. All stages share a reconstruction loss that combines $\ell_1$ and SSIM on magnitude images normalized by the target maximum:
\begin{equation}
    \mathcal{L}_{\mathrm{recon}} = \lambda_{\mathrm{img}} \|\bar{\mathbf{x}}_{\mathrm{pred}} - \bar{\mathbf{x}}_{\mathrm{target}}\|_1 + \lambda_{\mathrm{ssim}} \bigl(1 - \mathrm{SSIM}(\bar{\mathbf{x}}_{\mathrm{pred}}, \bar{\mathbf{x}}_{\mathrm{target}})\bigr).
    \label{eq:loss}
\end{equation}
In Stage~1, all U-Net cascades are trained with metal-corrupted data as both input and target ($\mathbf{y}_{\mathrm{metal}} \rightarrow \mathbf{x}_{\mathrm{metal}}$); here DC is constructive because input and target lie in the same corruption domain, yielding a stable backbone initialization. In Stage~2, the U-Net weights are frozen and only the DCR modules are optimized, switching the target to artifact-free images ($\mathbf{y}_{\mathrm{metal}} \rightarrow \mathbf{x}_{\mathrm{clean}}$) so that each DCR specializes in reversing the artifacts DC reintroduces at its cascade. In Stage~3, the backbone remains frozen while the PPO agent learns a sampling policy via the reward in Eq.~\ref{eq:reward} and the DCR modules are periodically fine-tuned on the collected episodes every $K$ policy updates, allowing artifact correction to adapt to the acquisition patterns the agent discovers. This decomposition avoids the gradient conflict between DC and the clean-target loss, and lets the sampling policy and per-cascade correction co-evolve on top of a stable reconstruction backbone.

\section{Experiments and Results}
\label{sec:experiments}

\subsection{Datasets}
\label{sec:datasets}

We evaluate OpenMASC on two datasets produced by the Data Generation module, each available in both Cartesian and radial formats.

\paragraph{autoPET (Simulation dataset)} Using the Data Generation module (Sec.~\ref{sec:data_generation}), we generate paired clean and metal-corrupted MRI volumes from 200 subjects in the autoPET CT/PET dataset~\cite{gatidis2023autopet}. Each volume contains 36 axial slices simulated with the TSE protocol, with edge slices outside the artifact-affected region excluded. Phantom segmentations and simulation outputs were reviewed by professional radiologists. We partition the subjects into 160 for training, 20 for validation, and 20 for testing with no overlap, yielding 5760 training, 720 validation, and 720 test slices. Cartesian $k$-space has dimensions $400 \times 200$ (readout $\times$ phase-encode). Radial $k$-space consists of 200 center-out spokes with 400 samples each, derived from the same complex images via MRI-NUFFT~\cite{comby2025mri}.

\paragraph{fastMRI (fastMRI + augment)} A separate test set is constructed from 800 knee volumes in fastMRI~\cite{zbontar2018fastmri,knoll2020fastmri} to assess cross-dataset generalization. Metal artifacts are synthesized using the method of Kwon et al.~\cite{kwon2018learning}, modeling off-resonance distortion from a simplified implant geometry~\cite{shi2017metallic} different from the training set, RF profile modulation, and Jacobian-based intensity warping. Random rotation ($\pm 45^{\circ}$) and translation ($\pm 80$ pixels) produce diverse artifact patterns. Radial $k$-space is generated identically to the autoPET procedure. This dataset differs from training data in both anatomy and implant model.

\subsection{Implementation Details}
\label{sec:implementation}

MA-VarNet uses $T{=}6$ cascades, each with a U-Net refinement, trajectory-specific DC, and a lightweight residual U-Net as DCR. The PPO agent has a shared convolutional encoder with separate actor and critic heads over 200 actions (phase-encode lines or spokes). Stages~1 and~2 train with AdamW (learning rate $10^{-4}$) at randomly sampled acceleration between $4\times$ and $8\times$, with 8 center readouts pre-acquired. Loss weights are $\lambda_{\mathrm{img}}{=}1.0$, $\lambda_{\mathrm{ssim}}{=}0.3$. Stage~3 uses PPO with learning rate $3{\times}10^{-4}$ and reward scaling $\alpha{=}100$; DCR modules are co-trained every $K{=}5$ policy updates. All experiments run on a single NVIDIA A6000 GPU.

Evaluation is performed at $4\times$ ($B{=}50$ readouts) and $8\times$ ($B{=}25$ readouts) acceleration, both with 8 center readouts pre-acquired. We report PSNR and SSIM on magnitude images normalized by the ground-truth maximum, excluding the implant region.

\subsection{Baselines}
\label{sec:baselines}

We compare against two reconstruction networks: E2E-VarNet~\cite{sriram2020end}, a 6-cascade variational network, and FI-VarNet~\cite{giannakopoulos2024accelerated}, a dual-path variant operating in feature and image domains. Both are trained in MAR mode on the same paired data as OpenMASC. For fixed sampling policies, we use variable-density random (VD) and equispaced on Cartesian, and random, golden angle (GA)~\cite{feng2014golden}, and equispaced on radial. For learned sampling policies, we evaluate DQN~\cite{mnih2015human} and SS-DDQN~\cite{pineda2020active}, both paired with FI-VarNet as the reconstruction backbone. All baselines share the same pre-acquired center readouts and are evaluated at $4\times$ and $8\times$ acceleration.

\begin{table*}[t]
\centering
\caption{Cartesian $k$-space results at $4\times$ and $8\times$ acceleration. Each method combines a sampling strategy with a reconstruction network. The simulation dataset is generated from autoPET via the Data Generation module, and fastMRI + augment uses a different implant model for cross-dataset evaluation. Best results in bold.}
\label{tab:cartesian}
\setlength{\tabcolsep}{4pt}
\renewcommand{\arraystretch}{1.15}
\resizebox{\textwidth}{!}{%
\begin{tabular}{l cc cc cc cc}
\toprule
\multirow{2}{*}{Method} & \multicolumn{2}{c}{Simulation 4$\times$} & \multicolumn{2}{c}{Simulation 8$\times$} & \multicolumn{2}{c}{fastMRI 4$\times$} & \multicolumn{2}{c}{fastMRI 8$\times$} \\
\cmidrule(lr){2-3} \cmidrule(lr){4-5} \cmidrule(lr){6-7} \cmidrule(lr){8-9}
 & SSIM$\uparrow$ & PSNR$\uparrow$ & SSIM$\uparrow$ & PSNR$\uparrow$ & SSIM$\uparrow$ & PSNR$\uparrow$ & SSIM$\uparrow$ & PSNR$\uparrow$ \\
\midrule
Rand (VD) + E2E-VarNet      & 0.756 $\pm$ 0.015 & 25.83 $\pm$ 1.75 & 0.747 $\pm$ 0.016 & 25.46 $\pm$ 1.69 & 0.749 $\pm$ 0.029 & 28.72 $\pm$ 1.33 & 0.679 $\pm$ 0.035 & 27.77 $\pm$ 1.29 \\
Equi + E2E-VarNet       & 0.746 $\pm$ 0.017 & 25.45 $\pm$ 1.72 & 0.716 $\pm$ 0.022 & 24.07 $\pm$ 1.57 & 0.727 $\pm$ 0.027 & 28.14 $\pm$ 1.28 & 0.651 $\pm$ 0.032 & 27.00 $\pm$ 1.37 \\
Rand (VD) + FI-VarNet   & 0.759 $\pm$ 0.014 & 26.01 $\pm$ 1.71 & 0.752 $\pm$ 0.015 & 25.67 $\pm$ 1.65 & 0.751 $\pm$ 0.029 & 28.89 $\pm$ 1.27 & 0.683 $\pm$ 0.035 & 27.97 $\pm$ 1.24 \\
Equi + FI-VarNet    & 0.750 $\pm$ 0.015 & 25.64 $\pm$ 1.68 & 0.721 $\pm$ 0.021 & 24.31 $\pm$ 1.55 & 0.731 $\pm$ 0.027 & 28.26 $\pm$ 1.27 & 0.656 $\pm$ 0.033 & 27.10 $\pm$ 1.35 \\
SS-DDQN + FI-VarNet       & 0.760 $\pm$ 0.014 & 26.01 $\pm$ 1.70 & 0.755 $\pm$ 0.014 & 25.81 $\pm$ 1.65 & 0.750 $\pm$ 0.028 & 28.70 $\pm$ 1.23 & 0.685 $\pm$ 0.036 & 28.08 $\pm$ 1.23 \\
DQN + FI-VarNet           & 0.760 $\pm$ 0.014 & 26.01 $\pm$ 1.70 & 0.755 $\pm$ 0.014 & 25.78 $\pm$ 1.64 & 0.750 $\pm$ 0.028 & 28.70 $\pm$ 1.23 & 0.689 $\pm$ 0.034 & 28.08 $\pm$ 1.24 \\
\midrule
\textbf{OpenMASC (ours)} & \textbf{0.777 $\pm$ 0.013} & \textbf{27.62 $\pm$ 1.51} & \textbf{0.773 $\pm$ 0.013} & \textbf{27.38 $\pm$ 1.45} & \textbf{0.761 $\pm$ 0.030} & \textbf{29.34 $\pm$ 1.35} & \textbf{0.696 $\pm$ 0.037} & \textbf{28.51 $\pm$ 1.29} \\
\bottomrule
\end{tabular}%
}
\end{table*}

\begin{figure*}[t]
    \centering
    \includegraphics[width=\textwidth]{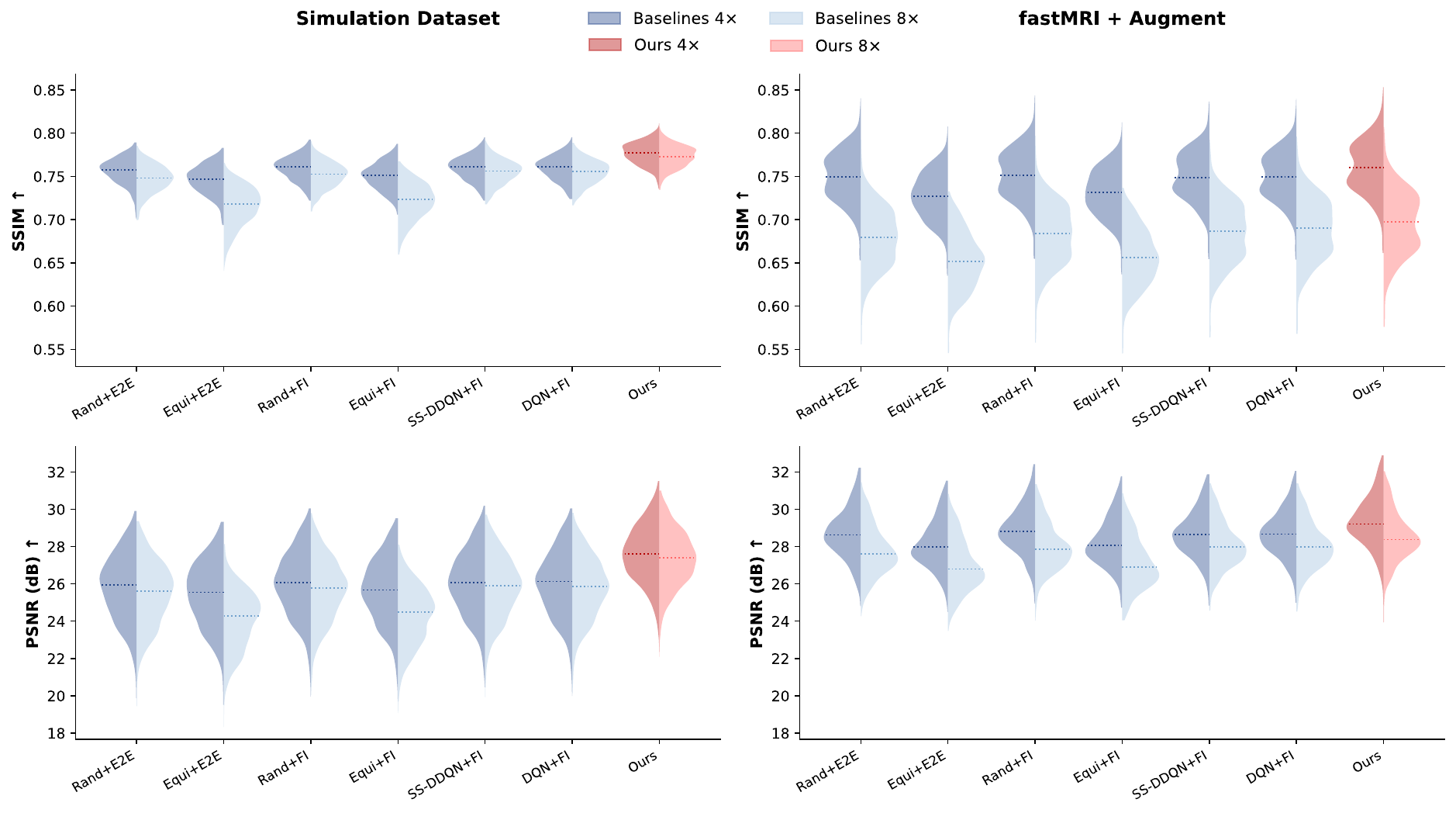}
    \caption{Cartesian $k$-space results shown as violin plots. Top row: SSIM; bottom row: PSNR. Left column: simulation dataset; right column: fastMRI + augment. Each method shows paired distributions for $4\times$ (blue) and $8\times$ (orange) acceleration. OpenMASC (red, rightmost) achieves the highest median and narrowest spread across all conditions.}
    \label{fig:radial_violin_cartesian}
\end{figure*}

\subsection{Cartesian Results}
\label{sec:cartesian_results}
 
Fig.~\ref{fig:radial_violin_cartesian} and Table~\ref{tab:cartesian} summarize Cartesian $k$-space results across both datasets and acceleration factors. OpenMASC achieves the highest SSIM and PSNR in every setting. On the simulation dataset at $4\times$, OpenMASC reaches an SSIM of 0.777 and PSNR of 27.62\,dB, compared to 0.760 SSIM and 26.01\,dB for SS-DDQN + FI-VarNet, the next best method. The improvement holds at $8\times$, where OpenMASC achieves 0.773 SSIM and 27.38\,dB versus 0.755 and 25.81\,dB for the same baseline. On fastMRI + augment, OpenMASC remains the top performer with 0.761 SSIM and 29.34\,dB at $4\times$, and 0.696 SSIM and 28.51\,dB at $8\times$. Among baselines, FI-VarNet consistently outperforms E2E-VarNet under the same sampling strategy, and the learned policies DQN and SS-DDQN provide modest gains over fixed sampling.
 
\begin{figure*}[t]
    \centering
    \includegraphics[width=\textwidth]{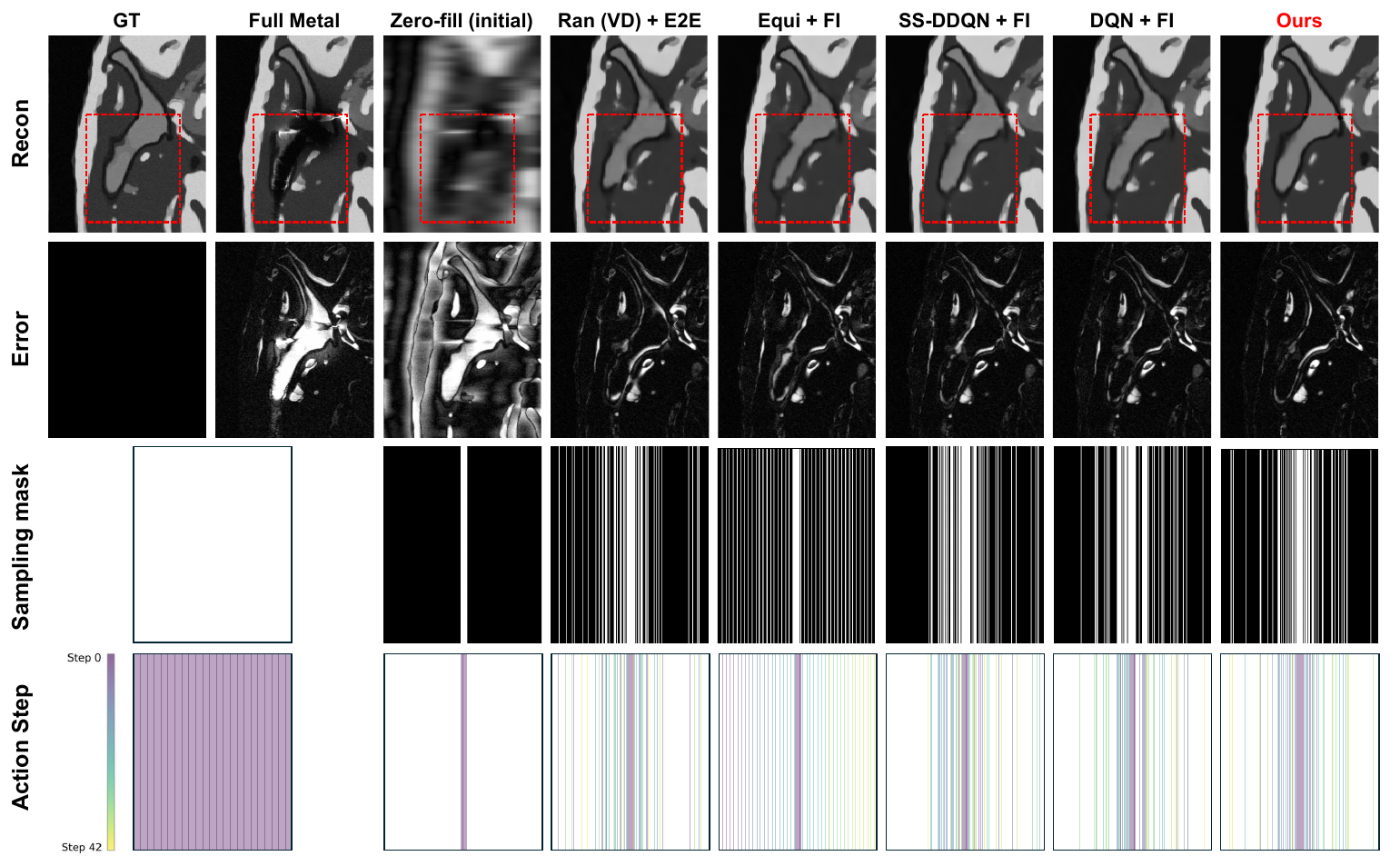}
    \caption{Qualitative Cartesian results on the simulation dataset at $4\times$ acceleration. Rows from top to bottom: reconstructed images with region of interest highlighted by red dashed boxes, error maps, sampling masks, and action step visualizations colored by acquisition order from Step~0 to Step~42. OpenMASC produces reconstructions closest to the ground truth with the darkest error maps.}
    \label{fig:cartesian_qual}
\end{figure*}

Fig.~\ref{fig:cartesian_qual} shows a qualitative comparison on the simulation dataset at $4\times$. Four rows display reconstructed images, error maps, sampling masks, and action steps colored by acquisition order. Baseline methods retain visible artifacts near the implant, highlighted by the red dashed box, whereas OpenMASC recovers finer structural detail with lower error.
 
\subsection{Radial Results}
\label{sec:radial_results}
 
\begin{figure*}[t]
    \centering
    \includegraphics[width=\textwidth]{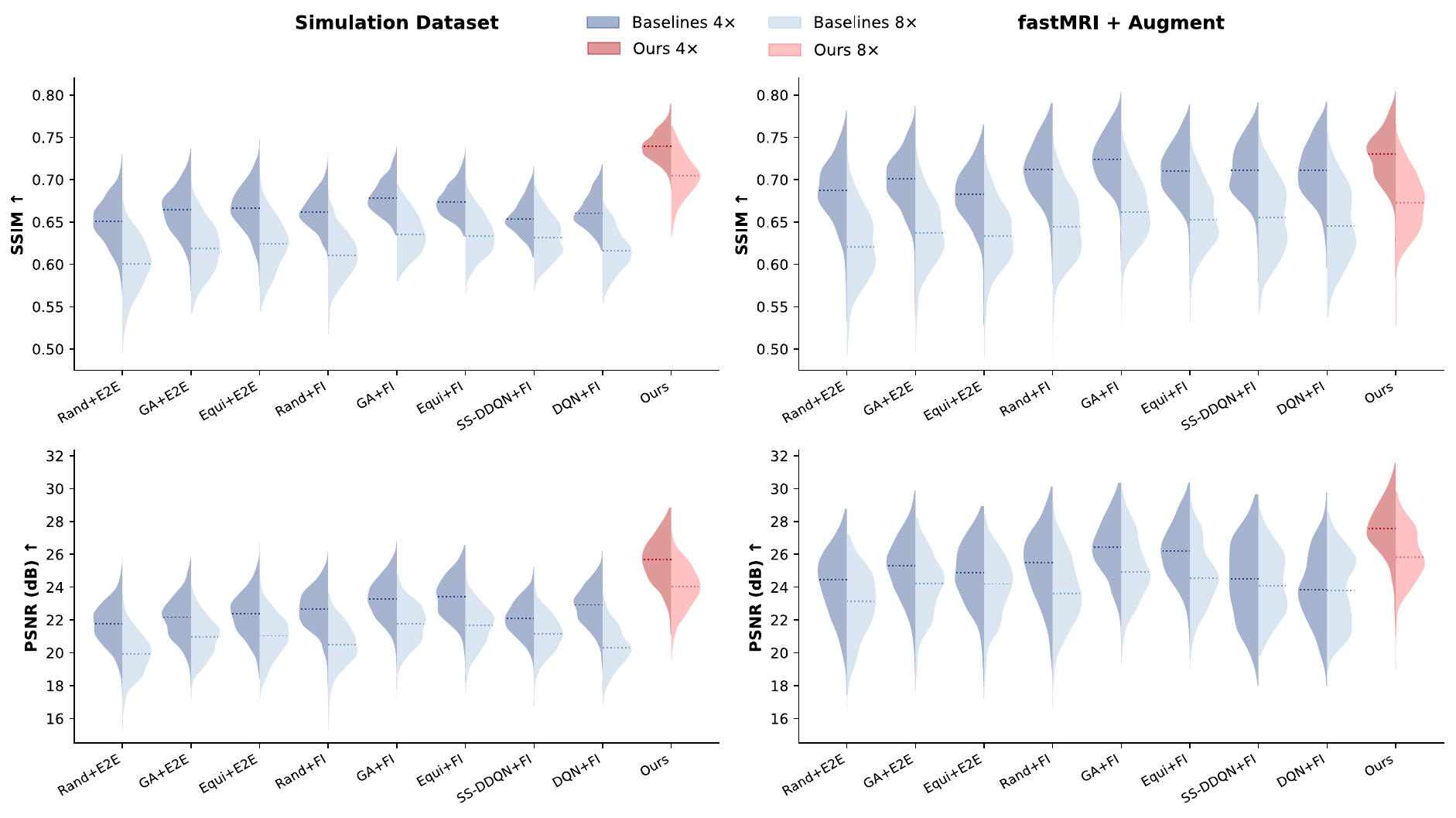}
    \caption{Radial $k$-space results shown as violin plots. Top row: SSIM; bottom row: PSNR. Left column: simulation dataset; right column: fastMRI + augment. Each method shows paired distributions for $4\times$ (blue) and $8\times$ (orange) acceleration. OpenMASC (red, rightmost) achieves the highest median and narrowest spread across all conditions.}
    \label{fig:radial_violin}
\end{figure*}

Fig.~\ref{fig:radial_violin} and Table~\ref{tab:results_radial} present radial $k$-space results, with the violin plots showing the per-slice metric distribution across the test set and the table reporting mean and standard deviation. The four panels display SSIM and PSNR for the simulation dataset and fastMRI + augment, at $4\times$ and $8\times$ acceleration.

OpenMASC achieves the highest median and tightest distribution in every condition. On the simulation dataset at $4\times$, OpenMASC reaches 0.740 SSIM and 25.61\,dB PSNR, compared to 0.680 SSIM and 23.26\,dB for golden angle + FI-VarNet. At $8\times$, the gap widens further, with OpenMASC at 0.704 SSIM and 23.96\,dB versus 0.636 and 21.71\,dB for the same baseline. On fastMRI + augment, OpenMASC maintains the best performance at both acceleration factors. The learned baselines SS-DDQN and DQN paired with FI-VarNet perform comparably to or below fixed golden-angle sampling on radial data.

\begin{figure*}[t]
    \centering
    \includegraphics[width=\textwidth]{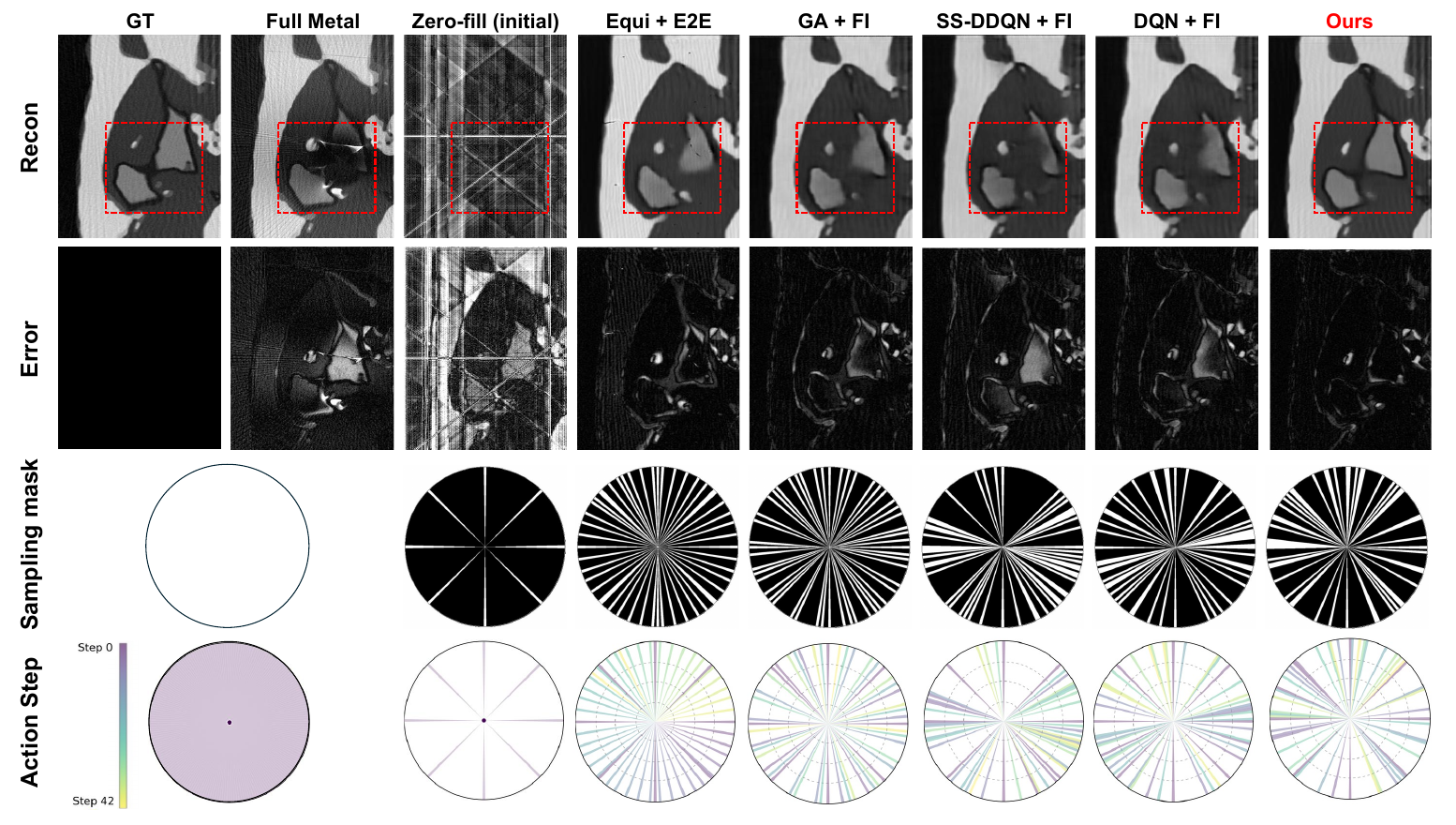}
    \caption{Qualitative radial results on the simulation dataset at $4\times$ acceleration. Rows from top to bottom: reconstructed images with region of interest (red dashed box), error maps, radial sampling masks, and action step visualizations colored by acquisition order. OpenMASC recovers sharp structural detail near the implant that baselines distort.}
    \label{fig:radial_qual}
\end{figure*}

Fig.~\ref{fig:radial_qual} shows qualitative radial results on the simulation dataset at $4\times$. Sampling masks and action steps are displayed as circular spoke patterns. Baselines exhibit streaking and blurred anatomy near the implant, while OpenMASC recovers sharper boundaries and tissue contrast.

\begin{table*}[t]
\centering
\caption{Radial $k$-space results at $4\times$ and $8\times$ acceleration. Each method combines a sampling strategy with a reconstruction network. The simulation dataset is generated from autoPET via the Data Generation module, and fastMRI + augment uses a different implant model for cross-dataset evaluation. Best results in bold.}
\label{tab:results_radial}
\setlength{\tabcolsep}{4pt}
\renewcommand{\arraystretch}{1.15}
\resizebox{\textwidth}{!}{%
\begin{tabular}{l cc cc cc cc}
\toprule
\multirow{2}{*}{Method} & \multicolumn{2}{c}{Simulation 4$\times$} & \multicolumn{2}{c}{Simulation 8$\times$} & \multicolumn{2}{c}{fastMRI 4$\times$} & \multicolumn{2}{c}{fastMRI 8$\times$} \\
\cmidrule(lr){2-3} \cmidrule(lr){4-5} \cmidrule(lr){6-7} \cmidrule(lr){8-9}
 & SSIM$\uparrow$ & PSNR$\uparrow$ & SSIM$\uparrow$ & PSNR$\uparrow$ & SSIM$\uparrow$ & PSNR$\uparrow$ & SSIM$\uparrow$ & PSNR$\uparrow$ \\
\midrule
Ran + E2E-VarNet         & 0.651 $\pm$ 0.027 & 21.73 $\pm$ 1.27 & 0.599 $\pm$ 0.032 & 19.86 $\pm$ 1.29 & 0.687 $\pm$ 0.038 & 24.23 $\pm$ 2.02 & 0.622 $\pm$ 0.042 & 22.95 $\pm$ 1.98 \\
GA + E2E-VarNet   & 0.663 $\pm$ 0.025 & 22.12 $\pm$ 1.27 & 0.619 $\pm$ 0.030 & 20.91 $\pm$ 1.21 & 0.698 $\pm$ 0.035 & 25.15 $\pm$ 1.90 & 0.638 $\pm$ 0.039 & 24.02 $\pm$ 1.88 \\
Equi + E2E-VarNet     & 0.667 $\pm$ 0.027 & 22.35 $\pm$ 1.30 & 0.623 $\pm$ 0.031 & 21.00 $\pm$ 1.31 & 0.681 $\pm$ 0.035 & 24.84 $\pm$ 1.94 & 0.633 $\pm$ 0.038 & 23.97 $\pm$ 1.91 \\
Rand + FI-VarNet          & 0.663 $\pm$ 0.020 & 22.63 $\pm$ 1.28 & 0.610 $\pm$ 0.027 & 20.44 $\pm$ 1.30 & 0.711 $\pm$ 0.034 & 25.34 $\pm$ 2.11 & 0.645 $\pm$ 0.040 & 23.70 $\pm$ 2.04 \\
GA + FI-VarNet    & 0.680 $\pm$ 0.019 & 23.26 $\pm$ 1.26 & 0.636 $\pm$ 0.024 & 21.71 $\pm$ 1.22 & 0.724 $\pm$ 0.031 & 26.42 $\pm$ 1.77 & 0.663 $\pm$ 0.037 & 25.09 $\pm$ 1.93 \\
Equi + FI-VarNet      & 0.676 $\pm$ 0.019 & 23.37 $\pm$ 1.28 & 0.635 $\pm$ 0.024 & 21.59 $\pm$ 1.25 & 0.710 $\pm$ 0.033 & 26.15 $\pm$ 1.75 & 0.655 $\pm$ 0.038 & 24.64 $\pm$ 1.89 \\
SS-DDQN + FI-VarNet         & 0.656 $\pm$ 0.019 & 22.07 $\pm$ 1.09 & 0.632 $\pm$ 0.025 & 21.11 $\pm$ 1.14 & 0.711 $\pm$ 0.033 & 24.39 $\pm$ 2.43 & 0.654 $\pm$ 0.042 & 24.17 $\pm$ 1.94 \\
DQN + FI-VarNet             & 0.663 $\pm$ 0.020 & 22.85 $\pm$ 1.21 & 0.616 $\pm$ 0.024 & 20.33 $\pm$ 1.20 & 0.711 $\pm$ 0.034 & 23.92 $\pm$ 2.17 & 0.647 $\pm$ 0.043 & 23.78 $\pm$ 2.14 \\
\midrule
\textbf{OpenMASC (ours)} & \textbf{0.740 $\pm$ 0.017} & \textbf{25.61 $\pm$ 1.41} & \textbf{0.704 $\pm$ 0.024} & \textbf{23.96 $\pm$ 1.33} & \textbf{0.727 $\pm$ 0.031} & \textbf{27.53 $\pm$ 1.66} & \textbf{0.672 $\pm$ 0.036} & \textbf{25.92 $\pm$ 1.60} \\
\bottomrule
\end{tabular}%
}
\end{table*}
 
\subsection{Ablation Studies}
\label{sec:ablation}
 
We isolate each component of OpenMASC on Cartesian autoPET at $4\times$. Table~\ref{tab:ablation} reports all results; statistical significance is assessed via paired $t$-tests against the full model ($p < 0.001$, $n{=}720$ slices).
 
Removing the DCR produces the largest degradation ($-6.2$\,dB PSNR). Among training ablations, skipping Stage~1 (VarNet pretraining) causes the biggest drop ($-0.5$\,dB), followed by removing the decoupled structure ($-0.15$\,dB) and Stage~3 co-training. For acquisition, replacing the learned PPO policy with equispaced sampling reduces SSIM by 0.009 (Cohen's $d{=}2.0$), and even against low-frequency biased random sampling the learned policy retains an advantage ($d{=}0.71$).
 
\begin{table}[t]
\centering
\small
\caption{Ablation study on Cartesian simulation dataset at $4\times$ acceleration. Each row removes one component from the full model. All paired $t$-tests vs.\ the full model yield $p < 0.001$ across $n{=}720$ slices. $^\dagger$Low-frequency biased random sampling. Best results in bold.}
\label{tab:ablation}
\setlength{\tabcolsep}{8pt}
\resizebox{\columnwidth}{!}{%
\begin{tabular}{l cc cc}
\toprule
Method
& SSIM $\uparrow$ & PSNR $\uparrow$
& MSE ($\times 10^{-3}$) $\downarrow$ & MAE ($\times 10^{-2}$) $\downarrow$ \\
\midrule
w/o Active Acq.\ (Rand$^\dagger$)
& 0.776 $\pm$ 0.013 & 27.53 $\pm$ 1.53
& 1.879 $\pm$ 0.691 & 2.312 $\pm$ 0.218 \\
w/o Active Acq.\ (Equi)
& 0.768 $\pm$ 0.013 & 27.09 $\pm$ 1.52
& 2.081 $\pm$ 0.754 & 2.429 $\pm$ 0.255 \\
w/o DCR
& 0.710 $\pm$ 0.022 & 21.45 $\pm$ 1.89
& 7.841 $\pm$ 3.365 & 3.716 $\pm$ 0.416 \\
w/o VarNet Pretrain
& 0.770 $\pm$ 0.013 & 27.11 $\pm$ 1.48
& 2.059 $\pm$ 0.724 & 2.417 $\pm$ 0.235 \\
w/o Decoupled Training
& 0.776 $\pm$ 0.013 & 27.47 $\pm$ 1.51
& 1.901 $\pm$ 0.674 & 2.318 $\pm$ 0.219 \\
w/o PPO-DCR Co-train
& 0.776 $\pm$ 0.013 & 27.55 $\pm$ 1.52
& 1.873 $\pm$ 0.687 & 2.303 $\pm$ 0.219 \\
\midrule
\textbf{Ours}
& \textbf{0.777 $\pm$ 0.013} & \textbf{27.62 $\pm$ 1.51}
& \textbf{1.837 $\pm$ 0.670} & \textbf{2.295 $\pm$ 0.213} \\
\bottomrule
\end{tabular}%
}
\end{table}

\section{Discussion}
\label{sec:discussion}

\subsection{Per-Cascade Correction and Co-Training}
\label{sec:disc_joint}

The ablation results in Table~\ref{tab:ablation} reveal that the DCR is the single most critical component, with its removal causing a 6.2\,dB drop in PSNR. This confirms that unrolled reconstruction networks, despite their strong performance on clean data, are fundamentally unsuited for metal-corrupted $k$-space without explicit correction at each cascade. The DC step, which is essential for convergence in standard accelerated MRI, becomes counterproductive when measurements carry susceptibility-induced phase errors. Each cascade reintroduces the artifacts that the preceding U-Net removed, and this accumulation cannot be undone by a single post-processing step at the output. The per-cascade DCR breaks this cycle by intervening immediately after every DC operation, before the corrupted signal can propagate further.

The benefit of joint optimization extends beyond the reconstruction architecture. Comparing the full model against configurations with frozen reconstruction in Table~\ref{tab:ablation}, the Stage~3 co-training of PPO and DCR yields measurable gains. When the acquisition agent and artifact correction are trained independently, each is optimized for a distribution the other does not produce: the agent selects readouts assuming a fixed reconstruction, while the reconstruction is trained on sampling patterns the agent never generates. Co-training resolves this mismatch by letting both components adapt to each other.

\subsection{Cross-Trajectory Generalization}
\label{sec:disc_trajectory}

A consistent finding across Table~\ref{tab:results_radial} and Fig.~\ref{fig:radial_violin} is that existing learned acquisition policies DQN and SS-DDQN, which were originally developed for Cartesian grids, fail to outperform fixed golden-angle sampling on radial data. These methods learn to exploit the regular structure of Cartesian $k$-space, such as the concentration of energy along low-frequency phase-encode lines, and this learned prior does not transfer to the angular sampling geometry of radial trajectories.

OpenMASC avoids this problem through its trajectory-agnostic design. The acquisition agent observes magnitude images and binary masks, not raw $k$-space, so its learned policy carries no implicit assumption about the sampling geometry. The DCR operates in the magnitude domain, which is independent of how $k$-space was acquired. Only the DC operator changes between configurations. This architectural separation means that the same agent and correction modules can be deployed on both Cartesian and radial data, which has practical value in clinical settings where the choice of trajectory depends on the patient and the imaging goal. For example, radial acquisition is often preferred near metal implants for its motion robustness, while Cartesian remains the default for routine scans. A single framework that supports both eliminates the need to maintain separate pipelines.

\subsection{Limitations and Future Work}
\label{sec:disc_limitations}

Several limitations should be noted. First, all experiments rely on simulated metal artifacts generated from physics-based models. While the simulation captures the dominant sources of corruption, including susceptibility-induced field distortion and signal dephasing, real clinical scans may contain additional effects such as RF shielding, eddy currents, and patient-specific geometry that the simulator does not model. Validation on clinical data with real metallic implants is an essential next step.

Second, the current framework operates in a single-coil setting. Clinical MRI scanners use multi-channel receiver coils, and extending OpenMASC to parallel imaging would require incorporating coil sensitivity maps into both the DC step and the acquisition formulation. Third, all experiments focus on hip implants in one anatomical region. Generalizing to other implant types with different susceptibility values, such as titanium or stainless steel, and to other body regions such as the spine or knee, would broaden the clinical applicability of the pipeline.

Fourth, the reconstruction is performed on 2D slices independently. Extending to 3D volumetric acquisition and reconstruction could exploit inter-slice correlations and is a natural direction for future development. Finally, the current framework supports Cartesian and radial trajectories. Other sampling geometries such as spiral, which is increasingly used in rapid imaging, could be incorporated by adding the corresponding forward model to the DC block while keeping all other components unchanged.

\section{Conclusion}
\label{sec:conclusion}
 
We presented OpenMASC, an open-source pipeline for metal-aware accelerated MRI that covers the full workflow from data generation to deployment. The pipeline comprises three modules: a Data Generation module that automatically produces paired clean and metal-corrupted MRI data from public CT datasets in both Cartesian and radial formats, MA-VarNet with per-cascade DC Rectifier for artifact-aware reconstruction, and a PPO-based active sampling agent that co-trains with the reconstruction network through a decoupled three-stage procedure. The entire framework is trajectory-agnostic except for the DC operator, allowing users to select between Cartesian and radial sampling with no architectural changes.
 
Experiments on two datasets at $4\times$ and $8\times$ acceleration show that OpenMASC outperforms both conventional and learned baselines across all configurations. Ablation studies confirm the necessity of each component, with the DCR providing the largest individual contribution. The code, pretrained models, and data generation pipeline are publicly available to support future research in metal-aware MRI reconstruction

\section{Acknowledgments}
This research was supported by NIH R01DK135597 (Huo), DoD HT9425-23-1-0003 (HCY), R01 EB031078 (Yan), R21 EB029639 (Yan), R03 EB034366 (Yan), S10 OD030389, and NIH NIDDK DK56942 (ABF). This work was also supported by Vanderbilt Seed Success Grant, Vanderbilt Discovery Grant, and VISE Seed Grant. This project was supported by The Leona M. and Harry B. Helmsley Charitable Trust grant G-1903-03793 and G-2103-05128. This research was also supported by NIH grants R01EB033385, R01DK132338, REB017230, R01MH125931 and NSF 2040462. We extend gratitude to NVIDIA for their support by means of the NVIDIA hardware grant. This works was also supported by NSF NAIRR Pilot Award NAIRR240055. This manuscript has been co-authored by ORNL, operated by UT-Battelle, LLC under Contract No. DE-AC05-00OR22725 with the U.S.Department of Energy.

This paper describes objective technical results and analysis. Any subjective views or opinions that might be expressed in the paper do not necessarily represent the views of the U.S. Department of Energy or the United States Government.  

\section{Declaration of generative AI and AI-assisted technologies in the writing process}

During the preparation of this work, the author(s) used ChatGPT5/Claude Opus4.6 in order to check for writing errors and perform appropriate editing and refinement. After using this tool, the author(s) reviewed and edited the content as needed and take(s) full responsibility for the content of the published article.

\bibliographystyle{elsarticle-num} 
\bibliography{main}

\end{document}